\documentclass[english]{article}
\usepackage{eucal}
\usepackage[utf8]{inputenc}
\usepackage[T1]{fontenc}
\usepackage{babel}
\usepackage{amsmath}
\usepackage{wrapfig}
\usepackage{float}
\usepackage{amsfonts}
\usepackage{csquotes}
\usepackage{amstext}
\usepackage{amssymb}
\usepackage{graphicx}
\usepackage{comment}
\usepackage{subcaption}
\usepackage{fancyhdr}
\usepackage{siunitx}
\usepackage{hyperref}
\usepackage{authblk}
\hypersetup{
    colorlinks=true,
    linkcolor=blue,
    filecolor=magenta,      
    urlcolor=cyan,
}
\usepackage{graphicx}

\usepackage[backend=biber,style=numeric,sorting=none]{biblatex}

\title{Direct evaluation of progression or regression of disease burden in brain metastatic disease with Deep Neuroevolution}
\author[1]{Joseph Stember\thanks{joestember@gmail.com}}
\author[1]{Robert Young\thanks{youngr@mskcc.org}}
\author[2]{Hrithwik Shalu\thanks{lucasprimesaiyan@gmail.com}}
\affil[1]{Department of Radiology, Memorial Sloan Kettering Cancer Center, NY, NY, 10065}
\affil[2]{Department of Aerospace Engineering, Indian Institute of Technology Madras, Chennai, India, 600036}

\begin{document}

\maketitle

\begin{abstract}

\indent \textit{Purpose} A core component of advancing cancer treatment research is assessing response to therapy. Doing so by hand, for example as per RECIST or RANO criteria, is tedious, time-consuming, and can miss important tumor response information; most notably, such criteria often exclude lesions, the non-target lesions, altogether. We wish to assess change in a holistic fashion that includes all lesions, obtaining simple, informative, and automated assessments of tumor progression or regression. Because genetic sub-types of cancer can be fairly specific and patient enrollment in therapy trials is often limited in number and accrual rate, we wish to make response assessments with small training sets. deep neuroevolution (DNE) can produce radiology artificial intelligence (AI) that performs well on small training sets. Following recent work in which we used DNE to train the parameters of a small convolutional neural network (CNN) for MRI sequence identification, we have now used a DNE parameter search to optimize a CNN that predicts progression versus regression of metastatic brain disease.

\indent \textit{Methods} We analyzed 50 pairs of MRI contrast-enhanced images as our training set. Half of these pairs, separated in time, qualified as disease progression, while the other 25 images constituted regression. We trained the parameters of a relatively small CNN via “mutations” that consisted of random CNN weight adjustments and mutation “fitness.” We then incorporated the best mutations into the next generation’s CNN, repeating this process for approximately 50,000 generations. We applied the CNNs to our training set, as well as a separate testing set with the same class balance of 25 progression and 25 regression images.

\indent \textit{Results} DNE achieved monotonic convergence to $100 \%$ training set accuracy. DNE also converged monotonically to 100 $\%$ testing set accuracy. 

\indent \textit{Conclusion} DNE can accurately classify brain-metastatic disease progression versus regression. Future work will extend the input from 2D image slices to full 3D volumes, and include the category of "no change." We believe that an approach such as our could ultimately provide a useful adjunct to RANO/RECIST assessment. 

\end{abstract}

\pagebreak

\section*{Introduction}

Artificial intelligence (AI) in radiology has traditionally focused on single imaging studies, for example detecting, localizing and/or measuring pathology such as masses, hemorrhage or vascular occlusion. In terms of brain cancer, AI has focused mostly on detection or measurement of primary lesions such as gliomas, oligodendrogliomas and glioblastomas, sometimes going a step further and predicting genomics or histopathology. A small amount of work has used AI to follow change of such lesions over time, but has focused on volumetric change. 

Metastatic brain tumors are much more common than primary central nervous system (CNS) lesions, and are increasingly prevalent as new treatments prolong life for those suffering from cancers such as melanoma, lung and breast cancer. The standard way to measure change for brain-metastatic disease is the RECIST or the RANO-BM approach. The latter, however, employs one-dimension measurements and focuses only on the few largest lesions. We would like more complete change evaluation of all brain metastases, which could do a better job of detecting for example mixed response.

Deep Neuroevolution (DNE) is a type of Deep Learning that is able make reliable conclusions based on small amounts of training data. That data efficiency should make it easier to adopt at individual institutions, with specialized smaller research data sets, and may render the approach applicable to patients with rare diseases. We sought here to introduce DNE-based classifier that inputs pairs of images from the same patient at two different time points and outputs disease response into one of two categories: progression or regression.

\section*{Methods}

\subsection*{Data collection.}

We obtained a waiver of informed consent from our Institutional Review Board for this retrospective study. 

Criteria for progression from an initial scan, “scan 1,” and a follow-up scan, “scan 2,” are an increase in size and/or number of enhancing lesions.  Similarly, we define disease regression for image pairs \{scan 1, scan 2\} as a decrease in size and/or number of lesions between scans 1 and scan 2. We displayed all images (both training and testing sets) without labels and in scrambled order to two neuroradiologists (twenty and three years of experience) for assessment per RANO-BM criteria including the minor response category from RANO-LGG \cite{wen2017response}. All fifty progression cases satisfied the RANO criteria of progression of disease. All fifty regression cases satisfied RANO criteria for either complete response (12/50), partial response (31/50), or minor response (7/50).

Although stereotactic radiosurgery, targeting individual lesions, has shown increasing use, WBRT remains an important treatment option. Patient usually undergo WBRT when their disease burden is high, with numerous lesions in multiple regions of the infratentorial and supratentorial brain and in cases of leptomeningeal disease. WBRT can result in significant shrinkage of metastatic lesions. 

Hence WBRT can serve as a sort of pivot point regarding disease evolution. One can in general expect disease progression preceding WBRT, and then regression following WBRT. This was consistent with our data set. Most commonly for progression cases, the patients had received WBRT after scan 2. In these cases, the disease had gotten worse up to a point at which the austere step of this treatment was needed. Most commonly for regression cases, the patients underwent WBRT between scans 1 and 2. Here, the disease burden had reached a level requiring the treatment, then disease burden decreased as detected on the follow-up, scan 2. All patients had undergone these second scans over 30 days after WBRT. 

We screened patient MRIs and selected scan pairs that maximized change, either progression or regression. The mean time intervals between scans 1 and 2 was nine months, ranging from 3 weeks to four years. Exclusion criteria included images excessively degraded by artifact, lack of clearly discernible disease or change of disease burden, an insufficient number of images or lack of T1 post-contrast images, a mixed disease response between scans, no clear temporal pattern, or no parenchymal metastases (i.e., only leptomeningeal disease, osseous/dural-based metastases, or scalp lesions).

\begin{figure}[H]
\hspace*{-0.2cm}  
\includegraphics[width=12cm]{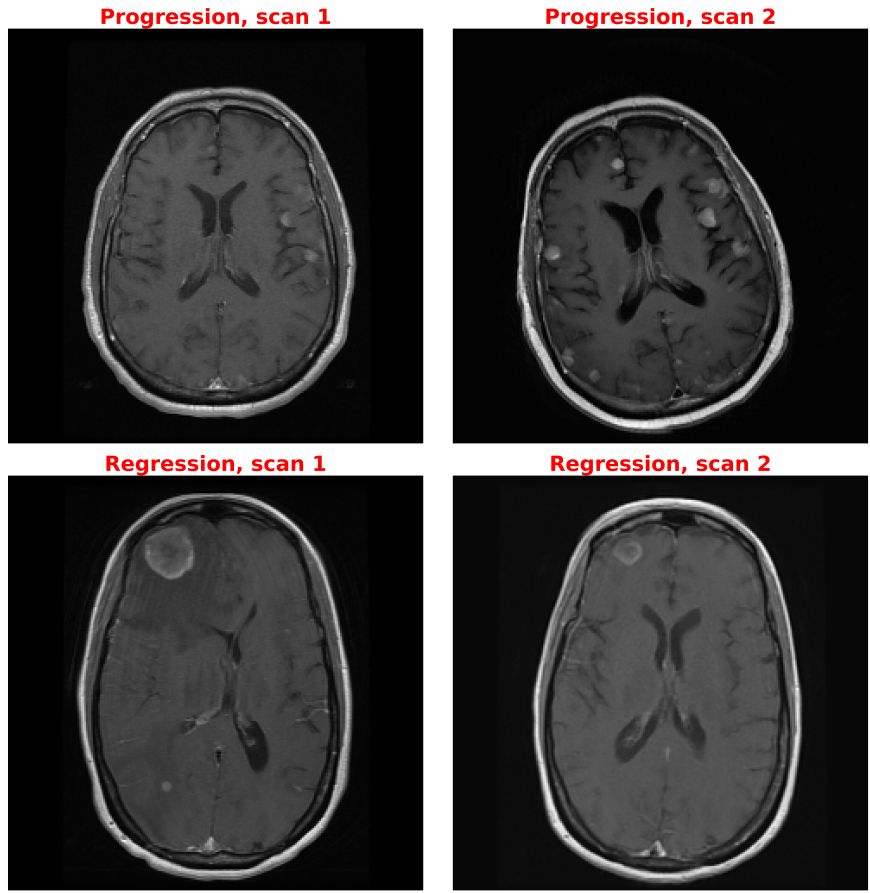}
\caption{Example of a progression case (top row) and a regression case (bottom row). For both cases, scan 1 refers to the first time point, and scan 2 refers to a subsequent time point.}
\label{fig:eg_images}
\end{figure}

Example cases of progression and regression are shown in \textbf{Figure} \textbf{\ref{fig:eg_images}}. The top row displays a progression case, in which the size and number of lesions increases in the time interval between scans 1 and 2. The bottom row of \textbf{Figure} \textbf{\ref{fig:eg_images}} displays an example of regression. Here, between scans 1 and 2, the dominant right frontal lesion decreases in size, and a small right parietal lesion resolves. Additionally, there is an improvement in mass effect, namely improved ventricular effacement and leftward midline shift. We note, however, that no increased mass effect is present for the top row/progression example. Interestingly, scan 2 from the progression case is tilted with respect to scan 1. However, we do not need to align images as prior researchers who used “classical machine learning” strategies did \cite{hajnal1995detection,freeborough1996accurate}.

As can be seen from \textbf{Figure 1}, we studied only post-contrast T1-weighted images, which display enhancing metastases. Most of the images were acquired from Memorial Sloan Kettering Cancer Center (MSK). We also used a few outside studies that had previously been scanned into our local PACS system for clinical review. Of the MSK-acquired scans: 
\begin{itemize}
    \item Most (85$\%$) scans had slice thickness of 5 mm, with a range of 2 mm to 7 mm.
    \item All but two scans were obtained with scanners from GE Medical Systems (Chicago, IL, USA). One each was acquired via scanners from Siemens (Munich, Germany) and one from Philips (Amsterdam, Netherlands).
    \item The contrast agent was Gadovist for all scans.
    \item The vast majority of scans (98$\%$) were obtained with a spin-echo sequence, the other 2$\%$ obtained via gradient echo sequence. All were 2D acquisitions.
    \item Mean repetition time (TR) was 1334 ms (range: 8-3266 ms).
    \item Mean echo time (TE) was 17 ms (range: 2-38 ms).
    \item Magnetic field strength was 1.5T in a slight majority (59$\%$) of cases, while the other 41$\%$ of scans were obtained with 3T.
    \item Pixel resolution ranged from 0.45 mm to 0.94 mm, with a mean resolution of 0.59 mm.
    \item Field of view (axial plane) ranged from 23 cm to 26 cm.
\end{itemize}
	
We selected pairs of 2D slices of our image volumes that displayed evidence of either progression or regression. We manually selected appropriate image pairs. Ultimately, this produced a training set of 50 image pairs, half of which represented progression, the other half being regression, again with corresponding coarse-grained RANO evaluations, as described above. Our training:testing split was 50:50 to ensure class balance, and we obtained a separate set of 50 testing images, also with class balance at 25 progression and 25 regression cases.

\subsection*{Convolutional neural network (CNN).}

\begin{figure}[H]
\hspace*{-0.3cm}  
\includegraphics[width=12.5cm]{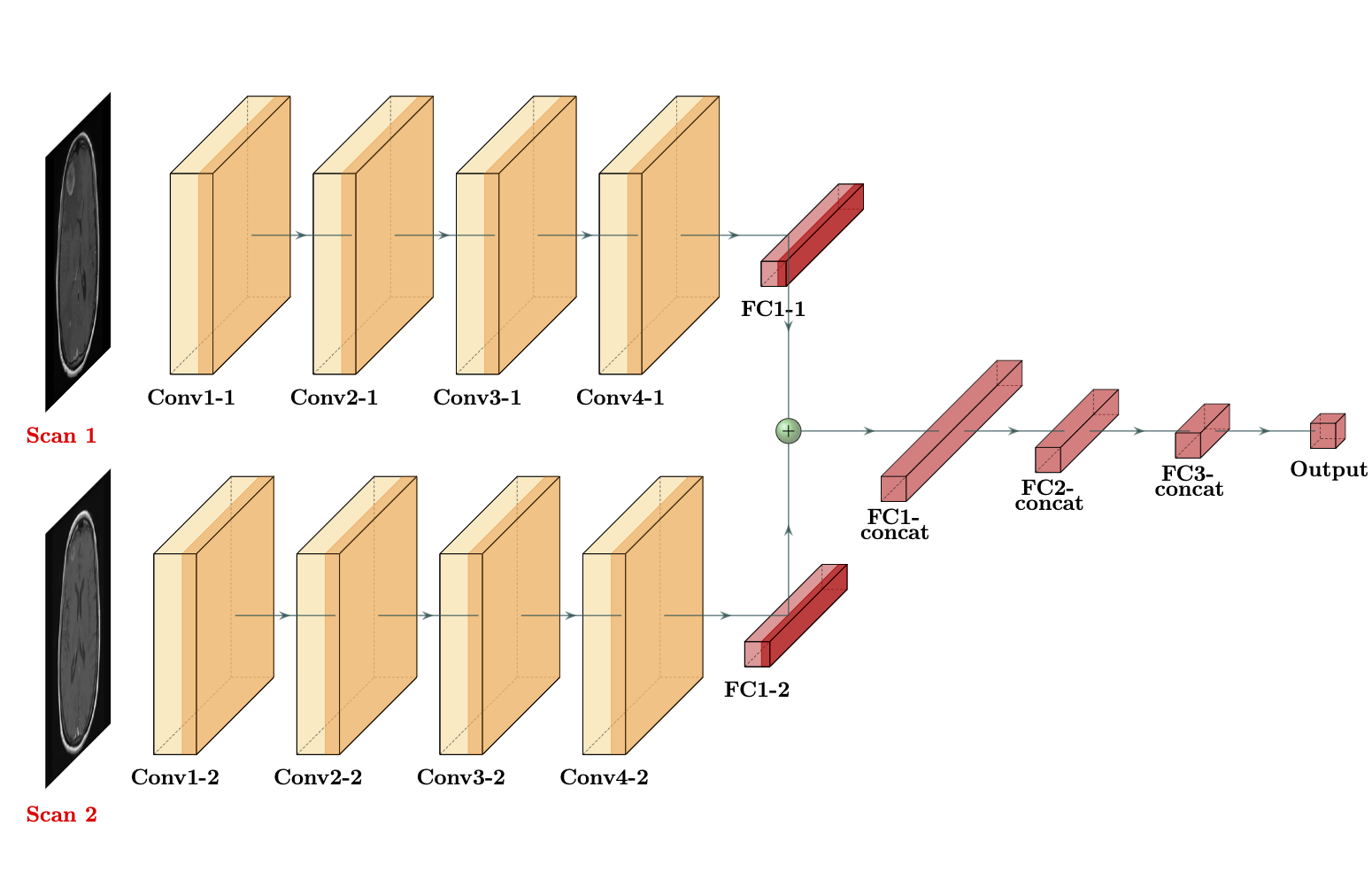}
\caption{CNN architecture illustrated with input images corresponding to the regression case from the bottom row of Figure \ref{fig:eg_images}}
\label{fig:CNN_architecture}
\end{figure}

We sought to train a classifier convolutional neural network (CNN). We used a relatively small CNN, because this permits faster clinical deployment. Because we were analyzing pairs of images, {scan 1,scan 2}, we input the two images into separate branches of the CNN. The CNN architecture is illustrated in \textbf{Figure} \textbf{\ref{fig:CNN_architecture}}.
Both branches consisted of four convolutional layers, each with 32 channels/feature maps, and followed by ReLu activation. The last convolutional layer was flattened and then passed to a fully connected layer of length 256 nodes for scans 1 and 2, named FC1-1 and FC1-2, respectively. FC1-1 and FC1-2 were then concatenated to form FC1-concat, of length 512. We then passed FC1-concat to a series of two fully connected layers, of length 256 and then 128 nodes. We performed SELU activation on each fully connected layer. We connected/passed the last fully connected layer, FC3-concat, to a two-node output layer. The two nodes represented the two classes of progression and regression. Upon a forward pass, the larger node determined the predicted class. We ded not apply an activation function to the output nodes, as the values of these two formed the CNN outputs directly.

We initialized all weights in our CNN via the standard Glorot randomization. We used kernels/filters composed of $3 \times 3$ weights. We employed a stride of two in the x and y directions, with padding of one applied to the input at each step.
Importantly, we did not use standard gradient descent to tune the CNN’s parameters. Hence, we required no optimizer or learning rate. We did not need to normalize weights or gradients via Dropout, Batch Normalization, or the Softmax function; gradients are not part of the DNE scheme. We can eschew the aforementioned techniques because we avert the problem of vanishing and exploding gradients by avoiding stochastic gradient descent.

\subsection*{Random mutations of CNN weights and fitness evaluation based on classification} accuracy.

Here we followed closely the Evolutionary Strategies approach from a prior proof-of-principle DNE work \cite{stember2021_DNE_sequence}. Interested readers can refer to the latter for a more detailed account of the approach, particularly as laid out in the Appendix. We briefly recapitulate the overall aspects below.

We began with a “first generation” lone parent CNN whose weights were randomly initialized as above. For each training set image, we produced 40 “children” CNNs with random offset mutations of the CNN weights \cite{stember2021_DNE_sequence}. We performed an inference/forward pass of each child CNN on all of the 50 training set image pairs. Each training set image pair received a reward of +1 if the child CNN correctly predicted the class, progression or regression, based on the label of zero or one provided. The total cumulative reward for a given CNN thus ranged in possible values from zero to fifty.

\subsection*{Deep neuroevolution selected for the fittest mutations and passed them on to future generations.}

We employed the total cumulative reward, i.e., classification accuracy, as our fitness criterion. As such, it guided the selection of the best mutations across generations.

We then incorporated the mutations of the better-performing child CNNs into the parent CNN’s “genome” of weights. We prioritized how much influence an incorporated mutation had in the parent’s genome by relative fitness level/accuracy of the mutated child CNN. The mutation-transformed parent then became the parent for the next generation. That parent would give rise to another generation of 40 children CNNs in exactly the same manner, via random mutations to the network weights. The selection and mutation incorporation process would repeat for many generations \cite{stember2021_DNE_sequence}. Restating a source of possible semantic and conceptual confusion in our analogy to biological evolution, here the children themselves did not directly propagate into the next generation. Rather, their best mutations propagated into the next generation’s via the new parent CNN \cite{stember2021_DNE_sequence}.

The above “many generations” amounted to a long training time required to produce high accuracy adaptations, as in prior DNE work \cite{stember2021_DNE_sequence}. Specifically, we trained for over 50,000 generations, taking roughly 25 hours of Google Colab CPU time. This is because, as in nature, most mutations are neutral at best or often harmful; only a small subset of mutations increase fitness. However, a benefit of evolutionary strategies is their known ability to converge monotonically toward the global optimum. We did achieve this convergence, as shown in \textbf{Figures} \textbf{\ref{fig:training}} and \textbf{\ref{fig:testing}} and discussed further below.



\section*{Results}

\begin{figure}[H]
\hspace*{-1.5cm}  
\includegraphics[width=15cm]{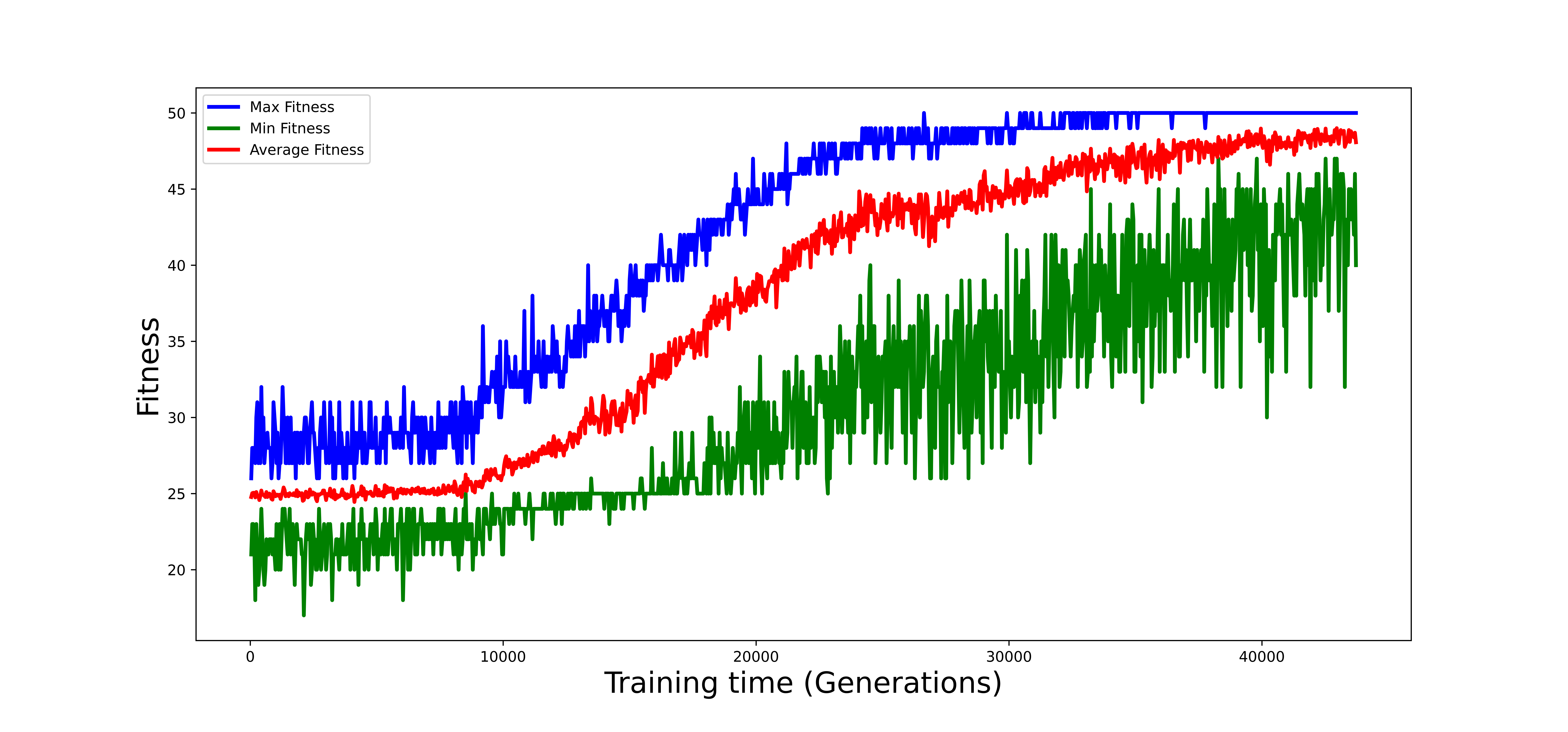}
\caption{Training set accuracy as a function of time in units of generations. The accuracies are for the children CNNs. The best-performing child (blue) is plotted along with the average of children (red), as well as the worst-performing child (green). These are ultimately all seen to converge on perfect accuracy (fitness of 80/80 correctly predicted image sequences.)}
\label{fig:training}
\end{figure}

The training time of DNE consisted of over 50,000 generations of evolution, after which global convergence was clearly manifest. The training time was 20 hours and 14 minutes on an 8-core CPU (no GPU used) in Google Colab Pro. Because we did not perform backpropagation of the CNN (no SGD), we did not require a GPU. We also did not perform any particular hyperparameter tuning at any point during training.

\begin{figure}[H]
\hspace*{-1.5cm}  
\centering
\includegraphics[width=15cm,height=8cm]{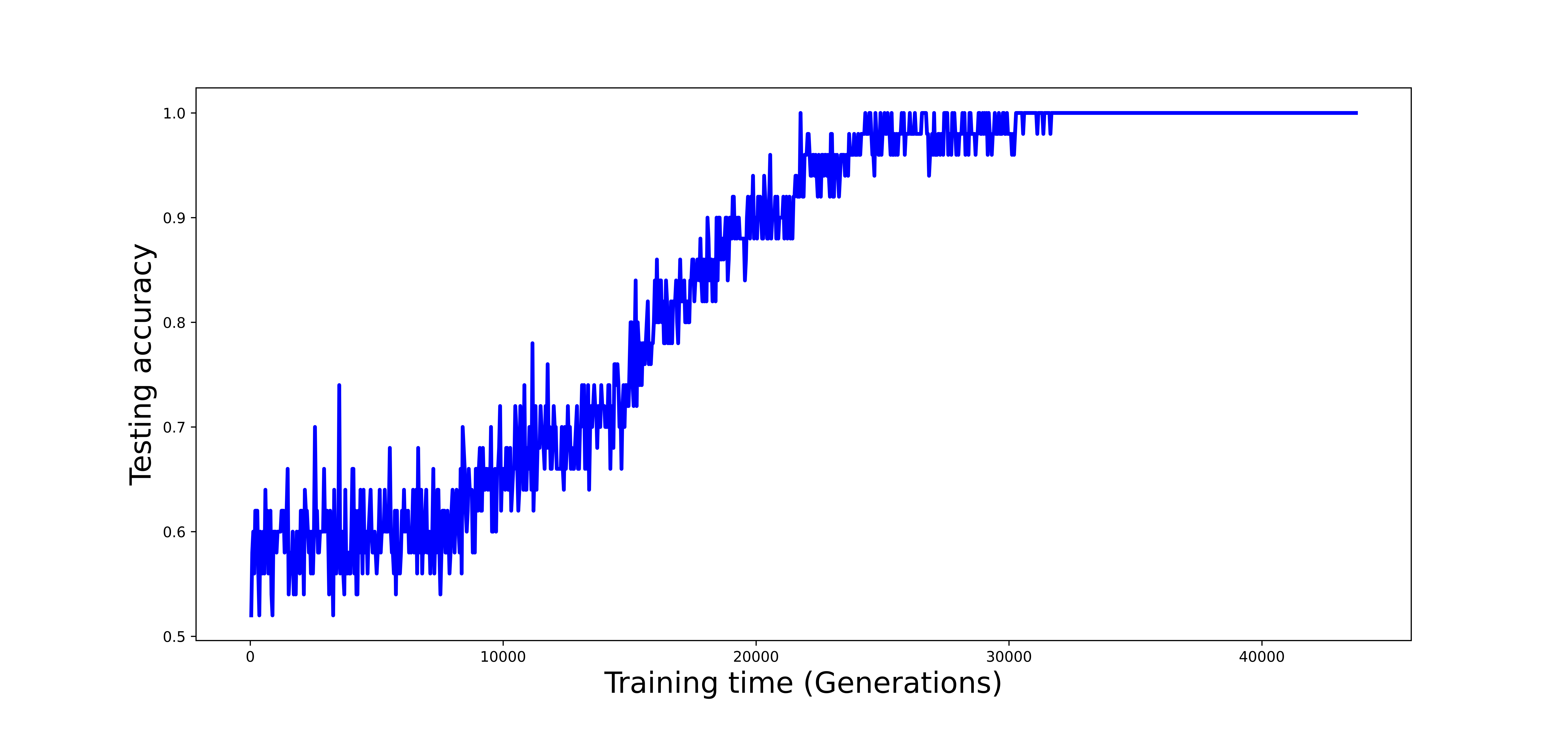}
\caption{ Testing set accuracy as a function of time in units of generations. Accuracy is evaluated as the maximum value over the 40 children CNNs and single parent CNN. We see monotonic convergence to $100 \%$ testing set accuracy}
\label{fig:testing}
\end{figure}

Training results for the child CNNs are shown in Figure \ref{fig:training}. As we saw in the prior work \cite{stember2021_DNE_sequence}, DNE converges strictly and monotonically to the global optimum, achieving 100$\%$ accuracy. After enough training, the average performance and even the lowest-performer "catch up" to the best performer and themselves converge toward the optimum. 

The application of the parent CNNs, which update at each generation by incorporating the best mutations/random weight offset vectors into their "DNA"/CNN weights, to the separate testing set is shown in Figure \ref{fig:testing}. As in our prior proof-of-principle work \cite{stember2021_DNE_sequence}, DNE achieves monotonic convergence of the testing set to $100 \%$ accuracy.  

\section*{Discussion}

\subsection*{Radiology AI classification often focuses on the detection of abnormalities on a single scan, but we often care about the difference between successive scans.}

AI-based classification has hitherto focused largely on classification that distinguishes on a single scan between normal and abnormal/disease states \cite{adhya2021positive,al2022impact,dehkharghani2021high}. While this is helpful to prioritize scans that need further urgent evaluation in the community and emergency room setting, most patients will have essentially normal scans. It is the exceptions that merit further examination and diagnosis by the radiologist. 
Many patients with chronic conditions or cancer, however, undergo serial scans throughout the course of their disease. Our chief motivation for the study is to diagnose clinically significant change over time. Radiologists manually inspect juxtaposed scans, including the current/most recent examination and prior comparison studies. Much of machine learning work to determine change has focused on volumetric change \cite{patriarche2004review,filipek1991volumetric,clarke1998mri} and/or subtraction of successive scans \cite{hajnal1995detection,freeborough1996accurate}. In the past decade, as radiology machine learning has largely moved on from “classical” techniques to AI and deep learning, the emphasis on volumetric comparison has continued \cite{kickingereder2019automated}.

\subsection*{The importance of monitoring change in brain-metastatic disease.}

Brain metastases from primary cancer outside of the central nervous system (CNS) are an important contributor to cancer morbidity and mortality. Brain metastases constitute the most common intracranial tumor type \cite{mut2012surgical}. An estimated 20-40$\%$  of all cancer patient will develop brain metastasis during the course of their illness \cite{tsao2005radiotherapeutic}. The prevalence continues to increase as new treatments allow for longer survival of patients with non-CNS systemic cancers, such as breast cancer, lung cancer, and melanoma \cite{national2018brain,uk2018brain}. In the past, treatment of metastatic brain disease emphasized palliation and often consisted of whole brain radiation therapy (WBRT) only. Currently, the focus has shifted toward maximizing survival, with a combined approach that employs surgery, stereotactic radiation therapy, and/or immunotherapy \cite{ahluwalia2014brain}.

Commensurately, care teams now place increased emphasis on follow-up of brain-metastatic disease, often over many years and many scans. Disease response, and hence its assessment, is crucial for determining which treatment changes, and when, are appropriate to maximize longevity and quality of life.

\subsection*{RANO and RECIST are the prevalent formal methods to assess treatment response.} 

A key component of advancing brain metastasis treatment research is the assessment of therapy response. Imaging typically largely drives this assessment. 

Most clinical trials employ the Response Evaluation Criteria in Solid Tumors (RECIST) 1.1 \cite{eisenhauer2009new} or Response Assessment in Neuro-Oncology brain metastases (RANO-BM) standardized response criteria \cite{lin2015response}. These criteria require measurement of the single largest diameter in 2 (RECIST 1.1) to 5 (RANO-BM) “target” brain metastases to determine complete response, partial response, stable disease, or progression. 

These parameters were selected to optimize a human reader’s reproducibility, time, and effort. While providing quantitative assessments of treatment response, they share the major limitation of not evaluating all brain metastases. Patients often have many more than five brain metastases (often up to 30, sometimes $\geq$ 50). 

Unsurprisingly, neglect of the vast majority of brain metastases may inadequately characterize overall tumor burden. The unmet clinical need is for an AI approach that directly classifies metastatic brain disease burden response to treatment, including response information about every metastasis, and can do so on small training sets, given the often small numbers of trial enrollees, especially at the early stages. 

\subsection*{Prior approaches to follow cancer burden response with artificial intelligence and the benefits of classification over size comparison.}

Recent work has used supervised deep learning to measure the volumes of primary brain tumors \cite{chang2019automatic,peng2022deep}, comparing their volumes across successive scans as a surrogate to RANO \cite{kickingereder2019automated}. In contrast to primary brain tumor patients who typically have just one target, patients with brain metastases often have many targets. These can be difficult for humans to count and keep track of, let alone trace each lesion’s contour. Tracing contours for all metastases, many of which are small and subtle, would be an overwhelming task. Hence, the annotation needed for a deep learning network would not be practical to obtain \cite{hsu2021automatic}.

Comparison of volumetric (or one/two-dimensional) measurements for brain metastases is suboptimal in other ways, as well. Minor volume prediction errors can profoundly affect percentage size change for smaller, more numerous metastatic lesions. Much of the prior research mentioned above has focused on primary gliomas, which tend to be larger. Hence, errors or variations in size or volume measurement have a more negligible effect on percentage change because the size/volume denominator is large. Primary brain tumors often consist of a single larger lesion, so errors do not typically add up, and are unlikely to produce a high \textit{percentage} error given the larger lesion's high volume denominator. In contrast, cumulative errors can compound for metastases, which are often sub-centimeter and numerous; the percentage errors are higher for each smaller lesion given their low volume denominators, and these higher percentage errors add up over the lesions.

\subsection*{Evolutionary strategies have a long history in machine learning.}

Evolutionary strategies, and the related field of genetic algorithms, are machine learning algorithms that are inspired by biological evolution. With a history going back to the 1960’s \cite{emmerich2018evolution,beyer2002evolution}, evolutionary strategies found early success primarily in engineering, where they often uncovered effective designs that had never before been considered by engineers \cite{emmerich2018evolution}. Evolutionary strategies have crucial and remarkable strengths and weaknesses. First the profound strength: these algorithms are known to explore optimization spaces thoroughly and inevitably arrive at global optima \cite{bertsekas2009convex}. Their chief weakness lies in the thorough and somewhat undirected exploration of parameters space that leads to the above convergence. Such a painstaking search results in long training times and/or the need for intensive computing capacity and parallelization.

\subsection*{Deep neuroevolution is a new sub-field of evolutionary strategies and genetic algorithms that shows promise for deep learning with small training sets and small networks.}

Deep neuroevolution (DNE) is a promising sub-field of evolutionary strategies/genetic algorithms that improves the performance of convolutional neural networks (CNNs) \cite{galvan2021neuroevolution}. CNNs, along with transformers, currently dominate deep learning applications in radiology. DNE computer vision research has focused mostly on hyperparameter tuning, both outside \cite{fernandes2021pruning,sun2019evolving,wang2021evolutionary} and inside of radiology \cite{ahmadian2021novel,hassanzadeh2021evolutionary,mortazi2018automatically}. DNE can also optimize neural network parameters, forming an alternative to gradient descent. DNE parameter optimization has found success in a variety of non-image-based tasks \cite{such2017deep,risi2019deep}, with early use in smaller artificial neural networks \cite{goerick1996evolution,montana1989training,porto1995alternative} and more recently with larger CNNs \cite{morse2016simple,zhang2017relationship}. Stember and Shalu used DNE to optimize the parameters of a CNN for MRI brain sequence classification, achieving global convergence with 100$\%$ testing set accuracy, despite a small training set of only 20 patients \cite{stember2021_DNE_sequence}.

Here we have applied DNE to the task of predicting \textit{change} between successive scans as a proof-of-principle. We chose the application of predicting disease progression versus regression in brain-metastatic cancer on MRI. This prediction was a \textit{direct classification task}, and was in particular not derived from differences in volumetric lesion predictions. 

\subsection*{Advantages of deep neuroevolution.}

In this study, we achieved perfect training and testing set accuracies using DNE with small training sets. No hyperparameter tuning was necessary. Because DNE is impervious to overfitting, we did not require any regularization devices such as Dropout, Batch Normalization, or Softmax function. For these reasons, DNE has the potential to provide the most reliable possible AI algorithms for deployment in clinical practice. 

\subsection*{Drawbacks, limitations, and future directions.}

One clear drawback with DNE is the sheer time required for training. This relatively simple task exceeded 24 hours. The promising upside is that the update process is highly parallelizable. We employed parallel processing with the threads available in the single CPU from Google Colab. However, we anticipate significant improvements in training time for future work with access to large numbers of CPUs.
Although we have emphasized the advantages of the mutation/selection-based parameter search of DNE, SGD does have some important advantages that might help to increase the speed of DNE training in future iterations. SGD provides directionality, sorely needed in DNE training. We anticipate that future iterations of DNE will incorporate SGD-derived directions to guide parameter search more efficiently.

Regarding our application of DNE to progression versus regression classification, we note that we studied only those two categories. Future work will include the category of “no significant change.” Additionally, the current study was performed on 2D slices, and future work will be generalized to the full 3D volumes.

We also anticipate the integration of DNE with deep reinforcement learning (DRL). Through DRL selection of intelligent actions correlated sequentially and with DNE optimization of DQN-predicted actions, we hope to master complex radiological control tasks that bring us closer to “meta” learning and true human-type intelligence in radiology.

Finally, we hope to validate the approach on non-MSK image data, where we could expect different scanner builds and settings, leading to the now-dreaded data drift problem of AI. Compile and validating on these outside data sets is a crucial and ongoing part of our work. However, we wish to note that the current work is a proof-of-concept, not yet a commercial grade industrial-ready algorithm.

\section*{Conflicts of interest}

The authors have pursued a provisional patent based largely on the work described here.

\section*{Funding}

We gratefully acknowledge external support from the Radiological Society of North America (RSNA) and the American Society of Neuroradiology (ASNR).

\printbibliography

\end{document}